\documentclass{article}
\usepackage{spconf,amsmath,graphicx}

\usepackage{enumitem}
\setlist{nosep, leftmargin=14pt}

\usepackage{mwe} 
\usepackage{multirow}

\usepackage{xcolor}
\definecolor{citecolor}{HTML}{0071BC}
\definecolor{linkcolor}{HTML}{ED1C24}
\usepackage[colorlinks,
            anchorcolor=red,
            citecolor=citecolor, 
            linkcolor=linkcolor,
            ]{hyperref}

\title{Leveraging AI Predicted and Expert Revised Annotations in Interactive Segmentation: Continual Tuning or Full Training?}
\name{Tiezheng Zhang $^1$ \qquad Xiaoxi Chen$^2$ \qquad Chongyu Qu$^1$ \qquad Alan Yuille$^1$ \qquad Zongwei Zhou$^{1,*}$}
\address{$^1$Johns Hopkins University \quad $^2$Shanghai Jiao Tong University\\
{\small \texttt{\textbf{Code \& Data:}} \href{https://github.com/ollie-ztz/Continue_Tuning}{\texttt{\textbf{https://github.com/ollie-ztz/Continue\_Tuning}}}}
}
\begin{document}
%
\maketitle
\begin{abstract}

Interactive segmentation, an integration of AI algorithms and human expertise, premises to improve the accuracy and efficiency of curating large-scale, detailed-annotated datasets in healthcare. Human experts revise the annotations predicted by AI, and in turn, AI improves its predictions by learning from these revised annotations. This interactive process continues to enhance the quality of annotations until no major revision is needed from experts. The key challenge is how to leverage \textit{AI predicted} and \textit{expert revised} annotations to iteratively improve the AI. Two problems arise: (1) The risk of catastrophic forgetting---the AI tends to forget the previously learned classes if it is only retrained using the expert revised classes. (2) Computational inefficiency when retraining the AI using both AI predicted and expert revised annotations; moreover, given the dominant AI predicted annotations in the dataset, the contribution of newly revised annotations---often account for a very small fraction---to the AI training remains marginal. This paper proposes \textit{Continual Tuning} to address the problems from two perspectives: network design and data reuse. Firstly, we design a shared network for all classes followed by class-specific networks dedicated to individual classes. To mitigate forgetting, we freeze the shared network for previously learned classes and only update the class-specific network for revised classes. Secondly, we reuse a small fraction of data with previous annotations to avoid over-computing. The selection of such data relies on the importance estimate of each data. The importance score is computed by combining the uncertainty and consistency of AI predictions. Our experiments demonstrate that Continual Tuning achieves a speed 16$\times$ greater than repeatedly training AI from scratch without compromising the performance.
\end{abstract}

\begin{figure*}[t]
\centering
\includegraphics[width=1.0\linewidth]{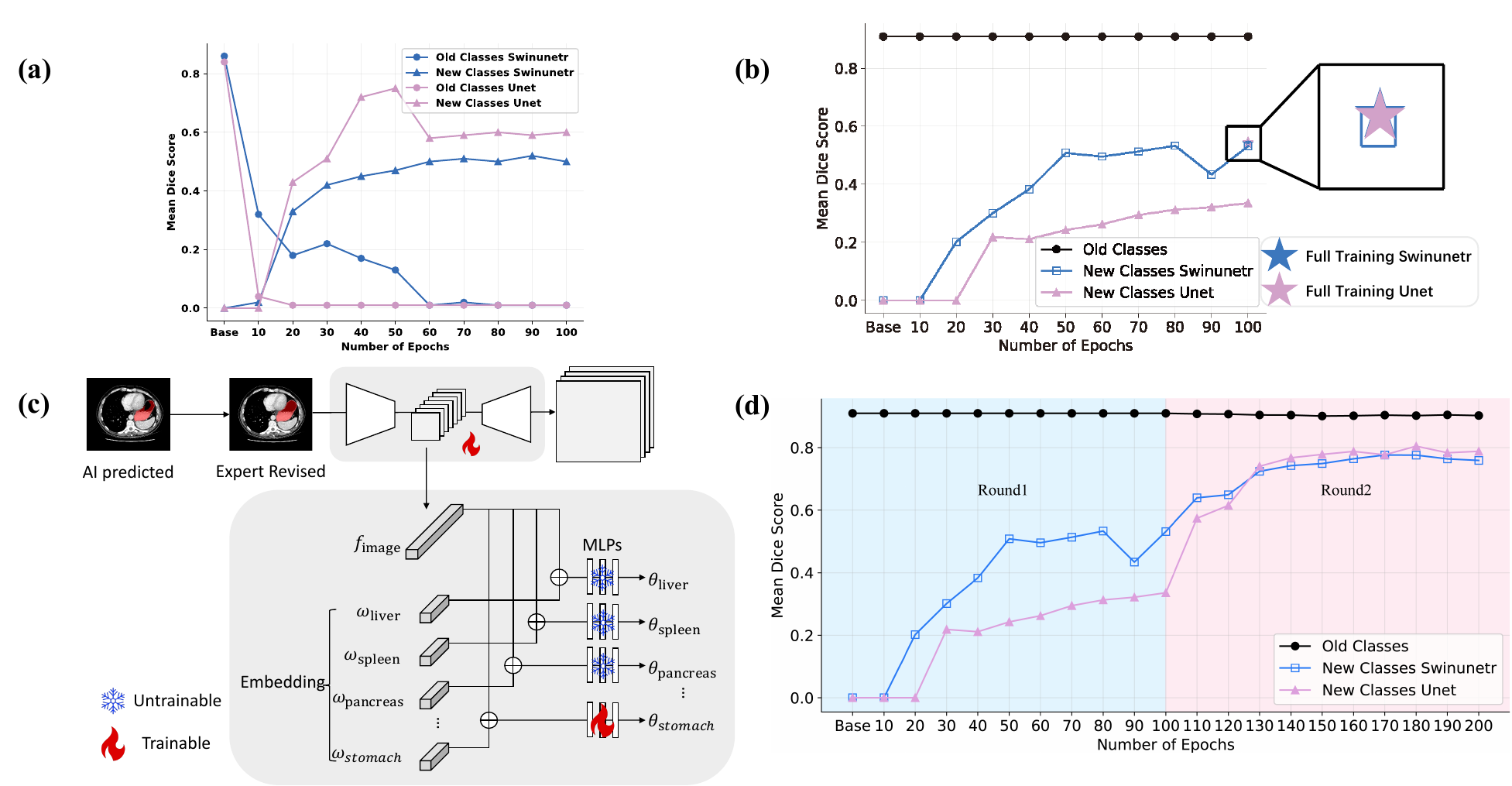}
    \caption{
    \textbf{(a) Catastrophic gorgetting in Swin UNETR~\cite{tang2022self} and U-Net~\cite{ronneberger2015u} backbones.} The old classes will be forgotten at the first few epochs when continual training AI models on data of new classes. \textbf{(b) Comparison of Continual Tuning and Full Training.
    } Two lines illustrate the mean DSC score using Continual Tuning method, while the asterisks show the final DSC score when applying Full Training. \textbf{(c) Shared Networks with Class-Specific Extensions.} The figure shows the networks we use, and we take the stomach as an example of the new class. 
    \textbf{(d) Results of Continual Tuning on Two Rounds.}
    The blue region represents first-round results of Continual Tuning, and the red region, the second-round results.
    }
\label{fig:structure}
\end{figure*}

\section{Introduction}
\label{sec:intro}

Combining AI algorithms with human expertise in interactive segmentation~\cite{olabarriaga1997setting,olabarriaga2001interaction,zhao2013overview} holds the promise of enhancing precision and productivity in the curation of large-scale, detailed annotated datasets such as SA-1B~\cite{kirillov2023segment}, TotalSegmentator~\cite{wasserthal2022totalsegmentator}, and AbdomenAtlas~\cite{qu2023annotating,li2024well}. During this synergy, human experts revise the AI predictions, and in return, AI enhances its predictions by adapting based on expert revised annotations. This iterative refinement continues until experts find that no substantial revisions are necessary~\cite{zhou2017fine,zhou2019integrating,zhou2021active,chen2023making,liu2021adaptive}.

However, the methodology to optimally leverage AI predicted annotations and expert revised annotations for the iterative enhancement of the AI remains ambiguous. There are two main issues to be considered. Firstly, there is the issue of catastrophic forgetting, which is shown in Figure~\ref{fig:structure}~(a), where the AI often overlooks previously learned classes if it is exclusively retrained on expert revised annotations. Secondly, the process of retraining the AI using both its predictions and expert revised annotations is not only computationally demanding but also less impactful. This is because the AI predictions largely dominate the dataset, making the contribution of expert revised annotations---often a small portion---almost negligible in the training process. 
In addressing the phenomenon of catastrophic forgetting~\cite{lewandowsky1995catastrophic}, one proposed strategy involves the retention of old class representations. For instance, Liu et al.~\cite{liu2022learning} advocate for the preservation of prototypical representations across diverse classes. Similarly, Lao et al.~\cite{lao2021two} employ a feature replay methodology.
Zhang et al.~\cite{zhang2023continual} use pseudo labels in their training process when the model is trained on new classes.
However, these methods, which depend on the accuracy of annotations, might encounter practical challenges. 
For example, inconsistent or incomplete annotations can lead to the creation of misleading classes or the replay of incorrect features.
Besides, Kirillov et al.~\cite{kirillov2023segment} proposed to retrain the AI from scratch, a method we referred to as \textit{Full Training}. However, this process could be time-consuming when applied to the medical domain. We seek to answer the following question: \textit{Can we utilize the AI predicted and expert revised annotations effectively in interactive segmentation?}
 
To answer this question, we propose \textit{Continual Tuning}, which focuses on two aspects: (i) network design and (ii) data reuse. \textbf{Firstly}, we develop a shared network that serves all classes, followed by different networks specifically designed for each class. To address the issue of forgetting, we keep the parameters of the shared network for the previously learned classes frozen while exclusively updating the network associated with the revised classes. As a result, the AI will not forget the previously learned classes, as shown Figure~\ref{fig:structure}~(b), while tuning only on the new classes revised by human experts. Additionally, Continual Tuning achieves a competitive Dice Similarity Coefficient (DSC) of 54.2\% and 16 $\times$ faster than Full Training. 
\textbf{Secondly}, we reuse a small fraction of data with previous annotations to avoid over-computing. The selection of such data relies on the importance estimate~\cite{qu2023annotating,li2013adaptive,gal2017deep} based on consistency, uncertainty, and overlapping. In summary, our ultimate goal is to continuously train AI models in interactive segmentation for better performance with the help of experts in the medical domain---this study makes a significant step towards it.

\section{Methodology}
\label{sec:method}

\begin{figure}[t]
    \begin{center}
    \includegraphics[width=1\columnwidth]{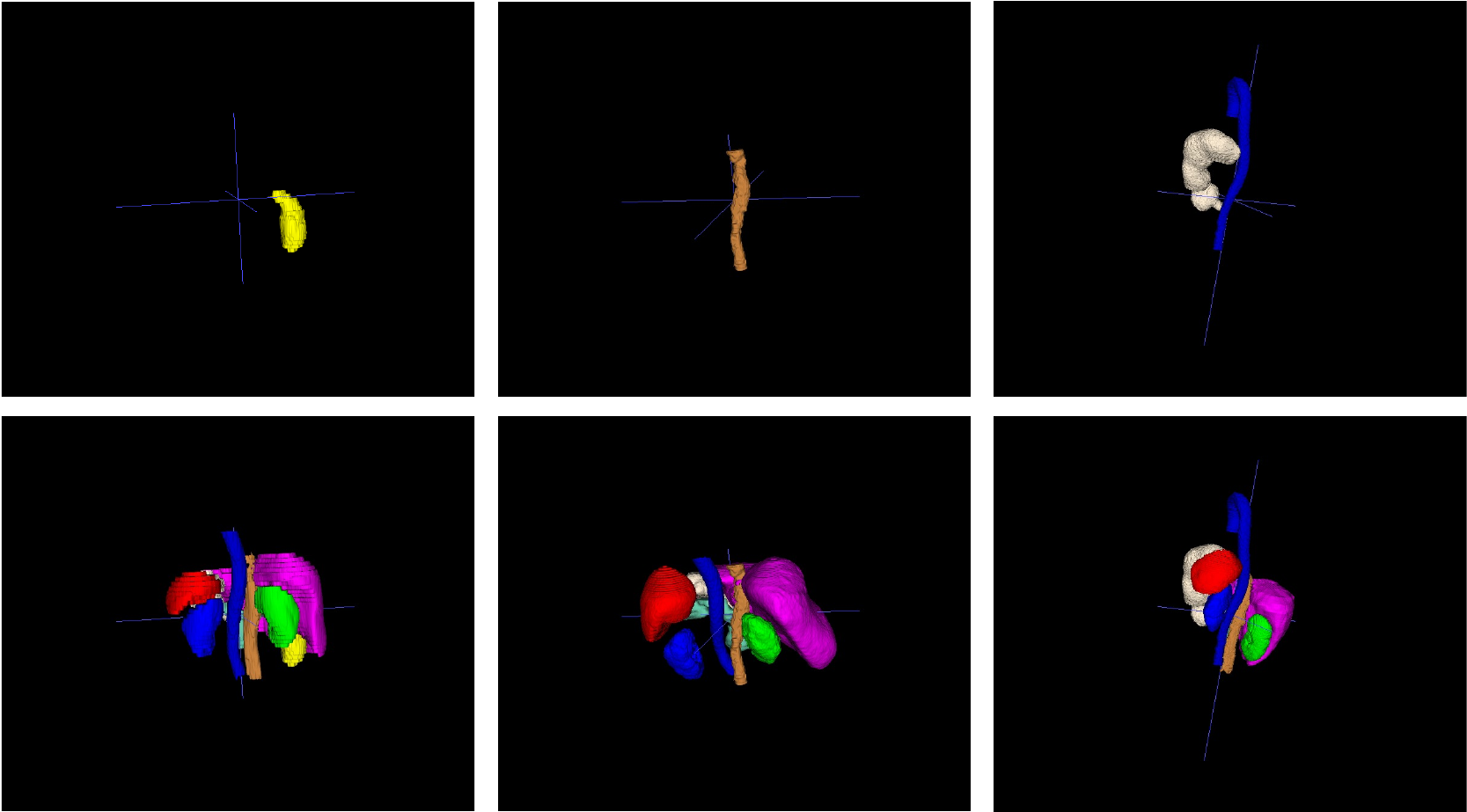}
    \caption{
    \textbf{
    Examples of Hybrid Data.
    }In the upper row, the revised annotations for gall bladder, postcava (IVC), and stomach \& aorta are presented from left to right. The lower row displays the corresponding hybrid annotations with old classes (liver, pancreas, left kidney, right kidney, and spleen). 
    }
    \label{fig:part_full}
  \end{center}
\end{figure}

\textbf{\textit{Continual Tuning}} ideally enables efficient refinement of AI models using revised annotations. For instance, AI models should enhance their aorta segmentation performance when solely fine-tuned on revised aorta annotations. Thus, we have devised a shared network architecture that operates in conjunction with networks tailored for specific classes, as illustrated in Figure~\ref{fig:structure}~(c). When fine-tuning AI models with expert revised annotations only, the shared network will remain unchanged, while the distinct networks associated with those revised annotations will be updated. With the help of text embeddings~\cite{liu2023clip}, which are encoded from the high-level visual semantics corresponding to each class, the class-specific networks become flexible to be updated.
For instance, as depicted in Figure~\ref{fig:structure}~(c), the AI models are fine-tuned exclusively with \textit{stomach} annotations, and only the networks corresponding to the stomach are updated. In general, given the CT scans with revised annotations ($X$), the parameters of the corresponding MLP layer could be updated with :
\begin{equation}
\begin{aligned}
    \theta_k = MLP_k(E(X),\omega_k)
    \label{formula:MLP_parameters}
\end{aligned}
\end{equation}
where $E(X)$ is the encoder feature of the image $X$, $\omega_k$ denotes the text embedding of each organ $k$.
From the perspective of the data itself, given adequate computational resources, one can train AI models from scratch utilizing both AI predicted and expert-revised annotations, referred to as Full Training~\cite{kirillov2023segment}. The improvement of the AI models could be slight due to the dominance of the unchanged annotations in the whole dataset. We propose to use expert-revised annotations in conjunction with AI predicted annotations (\textit{Hybrid Data Continual Tuning}) to achieve significant improvements in AI models beyond just slight enhancements. Specifically, we express AI predicted annotations for each CT scan as $(C_1, C_2, C_3, ... C_n)$ where $n$ is the total number of organs seen in this CT scan. The expert-revised annotations for each CT scan is $(C{^*}{_1}, C{^*}{_2}, C{^*}{_3}, ... C{^*}{_m})$ where $m$ is the number of organs revised by experts and $m\leq n$. By merging the revised annotations to the AI predicted annotations, the Hybrid Data could be expressed as$(C{^*}{_1}, C{^*}{_2}, C{^*}{_3}, ... C{^*}{_m},C_n)$ shown in Figure~\ref{fig:part_full}. 
This design enables AI models to efficiently prioritize expert revised annotations without forgetting, due to the use of AI-predicted annotations for previous classes.

\section{Experiments \& Results \& Discussion}
\label{sec:ERD}
To prove that our class-specific model and Hybrid Data continual tuning can effectively work in interactive segmentation, we proposed three experiment settings: one is focused on the model trained from one dataset, the other one is using the model trained from 14 publicly available datasets, another one is the comparison between the previous two.

\smallskip\noindent\textbf{Implementation Details.} 
The models were trained using the AdamW optimizer \cite{loshchilov2017decoupled}, coupled with a warm-up cosine scheduler lasting for 20 epochs \cite{goyal2017accurate}, and a weight decay of $1e^{-5}$. For the learning rate (\textit{lr}) and batch size, we opted for values of $1e^{-4}$ and 24, respectively. The pre-training phase extended over a total of 250 epochs. The training process was carried out across eight NVIDIA Quadro RTX 8000 cards.

\subsection{Continual Tuning models pre-trained on one dataset}
\label{sec:sub3.1}

We used randomly selected 200 CT scans with annotations from the AbdomenCT-1K dataset~\cite{ma2021abdomenct} to train AI models with Swin UNETR~\cite{tang2022self} and U-Net~\cite{ronneberger2015u} backbones. Those annotations comprise five classes: the liver, spleen, left kidney, right kidney, and pancreas. We asked an expert (over five years of experience) to annotate (using \href{https://aipair.com.cn/}{Pair}) four classes: the stomach, postcava, aorta, and gall bladder in 12 out of the 200 CT scans, which we refer to as the first round of expert revised annotations. By contrast, we also used the same CT scans with nine classes to train the model from scratch, referred to as Full Training. After fine-tuning these revised annotations, the AI models are used to infer another 200 CT scans from the AbdomenCT-1K dataset. Then, 22 out of the 200 CT scans selected for revision in four classes (stomach, postcava, aorta, and gall bladder) are used for continual fine-tuning, referred to as the second round. The selection for the revision process is based on the uncertainty of the AI predicted annotations.
To assess the performance of the models, we computed the DSC score on our proprietary JHH dataset containing high-quality annotations of all nine classes used in this experiment.

\smallskip\noindent\textbf{Results and Analysis.} The quantitative results in Figure~\ref{fig:structure}~(b) demonstrate that applying Continual Tuning on AI models could be 16$\times$ faster (200/12) compared with applying the Full Training method while still maintaining a similar DSC score (54.2\% vs. 54.4\%). The results in Figure~\ref{fig:structure}~(d) further demonstrate the promise of Continual Tuning in interactive segmentation tasks. The first round in the blue region indicates that Continual Tuning assists in preventing the issue of forgetting. Then, the sharp increase from the first round part to the second round in red regions is attributed to the 22 CT scans predicted by the AI models after the first round of learning. There might be more prevalent errors in these 22 CT scans, which, when revised by the experts, can further enhance the model's performance. The final average DSC scores can achieve about 76.1\% and 78.8\% for Swin UNETR and U-Net backbones, respectively. We expect AI model performance to improve gradually through interactive segmentation and Continual Tuning.

\begin{table}[t]
\scriptsize
\centering
\begin{tabular}{l|cc}
\multicolumn{1}{c|}{\textbf{Organ}}    & \textbf{\begin{tabular}[c]{@{}c@{}}Before Fine-tuning\\ mDice\end{tabular}} & \textbf{\begin{tabular}[c]{@{}c@{}}Revised Data Only \\Continual Tuning\\ mDice\end{tabular}} \\ \hline
{Spleen}           & 0.94                                                                            & 0.25                                                                           \\
{Right Kidney}     & 0.92                                                                            & 0.08                                                                           \\
{Left Kidney}      & 0.91                                                                            & 0.12                                                                           \\
{Pancreas}         & 0.81                                                                            & 0.07                                                                           \\
{Liver}            & 0.96                                                                            & 0.01                                                                           \\
{Stomach (11)}      & 0.93                                                                            & 0.90                                                                           \\
{Aorta (12)}        & 0.73                                                                            & 0.83                                                                           \\
{Postcava (IVC) (6)} & 0.76                                                                            & 0.75                                                                           \\
{Gall Bladder (1)}  & 0.82                                                                            & 0.82                                                                           \\ \hline
\multicolumn{1}{c|}{\textbf{Organ}}             & \textbf{\begin{tabular}[c]{@{}c@{}}Hybrid Data \\ Continual Tuning\\ mDice\end{tabular}}        & \textbf{\begin{tabular}[c]{@{}c@{}}Full Training\\ mDice\end{tabular}}          \\ \hline
Spleen                                  & 0.95                                                                            & 0.94                                                                           \\
Right Kidney                            & 0.92                                                                            & 0.92                                                                           \\
Left Kidney                             & 0.91                                                                            & 0.91                                                                           \\
Pancreas                                & 0.82                                                                            & 0.82                                                                           \\
Liver                                   & 0.96                                                                            & 0.96                                                                           \\
{Stomach (11)}      & 0.93                                                                            & 0.93                                                                           \\
{Aorta (12)}        & 0.83                                                                            & 0.75                                                                           \\
{Postcava(IVC) (6)} & 0.77                                                                            & 0.77                                                                           \\
{Gall Bladder (1)}  & 0.82                                                                            & 0.82                                                                          
\end{tabular}
    \caption{
The numbers in parentheses indicate the amount of revised CT scans. The table illustrates the mean DSC score obtained from implementing various data strategies on the AI model that has been trained using 14 datasets.}
\label{tab:liu_model_table_reply}
\end{table}
\subsection{Continual Tuning models pre-trained on 14 datasets}
\label{sec:sub3.2}
We first used 3,410 CT scans with annotations from 14 publicly available datasets to train the AI model with Swin UNETR backbones, which we refer to as the first round for this experiment. Those datasets are partially annotated but totally contain all nine classes used in the previous experiment. The AI model is used to infer another random 200 CT scans from the testing sets of 14 public datasets. Then, 12 out of the 200 CT scans selected for revision in four classes (stomach, postcava, aorta, and gall bladder) are used for continual fine-tuning, referred to as the second round. In this round, we tried to use three data strategies: one using revised annotations of 12 CT scans, the other one using 12 CT scans with nine classes, which is our Hybrid Data Continual Tuning method, and the last one is using all 200 CT scans with all nine classes.

\smallskip\noindent\textbf{Results and Analysis.} One difference between this experiment and the previous is the scale of datasets used to train the model. We hypothesize that this kind of model is closer to the model used in real scenarios. From Table~\ref{tab:liu_model_table_reply}, we could find that if the model is only fine-tuned with the revised CT scans, the model indeed improves the ability to segment the revised classes but also suffers from forgetting problems. Compared to using 200 CT scans in the second, using 12 CT scans could achieve a similar or better improvement of the model's ability. For example, the mean DSC score of the aorta improves by $\textbf{10\%}$ using Hybrid Data Continual Tuning, while it only improves by $\textbf{2}\%$ if we fine-tune the model with all 200 CT scans. This slight improvement is due to the dominance of unchanged data in the dataset (188 vs. 12).

\subsection{Continual Tuning: Impact on Model Scales}
\label{sec:sub3.3}

We used 200 CT scans from one dataset with nine classes to train AI models with Swin UNETR and U-Net backbones. The AI models are used to infer another random 200 CT scans from the testing sets of this dataset. Then the same amount of CT scans are selected for revision. And we applied the same data strategies as we did in~\S\ref{sec:sub3.2}. 

\smallskip\noindent\textbf{Results and Analysis.} From Table~\ref{tab:our_model_table_reply}, we could find that the models have better performance using all 200 CT scans, especially for organs that have revised annotations. Although the unchanged data still dominates the whole dataset, it does not weaken the influence of 12 revised annotations. The variations in phenotypes between~\S\ref{sec:sub3.2} and~\S\ref{sec:sub3.3} could be attributed to differences in dataset utilization. Multiple datasets could have different annotation principles. For example, some datasets include annotations for the stomach, including the cavity, while others may not. Although the annotation could be more accurate with the process of revision,  the model's performance could be minimized by different annotation principles. On the other hand, if the models are trained on a single dataset, each organ follows a consistent annotation principle, and more data could lead to better performance.

\begin{table}[t]
\centering
\scriptsize
\begin{tabular}{c|l|cc}
\textbf{Structures}          & \multicolumn{1}{c|}{\textbf{Organ}}     & \textbf{\begin{tabular}[c]{@{}c@{}}Hybrid Data \\Continual Tuning \\mDice\end{tabular}} & \textbf{\begin{tabular}[c]{@{}c@{}}Full Training\\ mDice\end{tabular}} \\ \hline
                             & {Spleen} & 0.93 & 0.93 \\
                             & {Right Kidney} & 0.92 & 0.92 \\
                             & {Left Kidney} & 0.90 & 0.90 \\
                             & {Pancreas} & 0.73 & 0.80 \\
                             & {Liver}  & 0.95 & 0.96 \\
                             & {Stomach (11)} & 0.77 & 0.89 \\
                             & {Aorta (12)} & 0.69 & 0.80 \\
                             & {Postcava(IVC) (6)} & 0.58 & 0.76 \\
\multirow{-9}{*}{Swin UNETR} & {Gall Bladder (1)}  & 0.43 & 0.82 \\ \hline
                             & Spleen & 0.92 & 0.90 \\
                             & Right Kidney & 0.90 & 0.92 \\
                             & Left Kidney & 0.89 & 0.90 \\
                             & Pancreas & 0.79 & 0.81 \\
                             & Liver & 0.95 & 0.95 \\
                             & {Stomach (11)} & 0.64 & 0.86 \\
                             & {Aorta (12)} & 0.70 & 0.79 \\
                             & {Postcava(IVC) (6)} & 0.55 & 0.76 \\
\multirow{-9}{*}{U-Net}       & {Gall Bladder (1)}  & 0.42 & 0.83 
\end{tabular}
    \caption{The numbers in parentheses indicate the amount of revised CT scans. The table illustrates the mean DSC score obtained from implementing various data strategies on the AI models that have been trained using one dataset.}
    \label{tab:our_model_table_reply}
\end{table}

\section{Conclusion}
\label{sec:Conclusion}

In this paper, we propose Continual Tuning that integrates network design and data reuse to leverage AI predicted and expert revised annotations during the interactive segmentation procedure. Continual Tuning enables AI models to be fine-tuned efficiently (16$\times$ faster in our experiment) only with expert revised annotations in interactive segmentation tasks in the medical domain. This reveals the great potential for fine-tuning the AI models with incoming partial class datasets, e.g., AbdomenCT-1K, or datasets containing tumors.

\smallskip\noindent\textbf{Clinical Application.} Our proposed Continual Tuning enhances diagnostic accuracy and minimizes annotation efforts, thus facilitating long-term learning and promoting trust in the model's decision-making process. This approach fosters continual improvement and the integration of the latest medical knowledge, thereby increasing the model's value in evidence-based healthcare settings.

\smallskip\noindent\textbf{Limitation.}
Continual Tuning involves several procedures that require human intervention, such as the annotation revision and selection process. This human involvement introduces a degree of subjectivity and variability, which may impact the overall quality and consistency of the annotations, consequently affecting the performance of the AI models. Secondly, the class-specific network we employ to prevent catastrophic forgetting is not inherently adaptive. As datasets evolve and new classes are introduced, the pre-defined class-specific network may become less effective.

\smallskip
\noindent\textbf{Compliance with Ethical Standards.}
Committee/IRB of Johns Hopkins Medicine gave ethical approval for this work.

\noindent\textbf{Acknowledgments.}
This work was supported by the Lustgarten Foundation for Pancreatic Cancer Research and the McGovern Foundation.
We thank Yaoyao Liu, Ju He for their constructive suggestions at several stages of the project.

{\small
\bibliographystyle{IEEEbib}
\bibliography{zzhou,refs}
}

\end{document}